\documentclass[letterpaper,10pt,conference]{ieeeconf}

\IEEEoverridecommandlockouts
\usepackage{amsmath,amssymb}
\usepackage{graphicx}
\usepackage[caption=false,font=footnotesize]{subfig}
\usepackage{cite}
\usepackage{url}
\usepackage{algorithm}
\usepackage{algpseudocode}
\newtheorem{lemma}{Lemma}
\newtheorem{theorem}{Theorem}

\newtheorem{assumption}{Assumption}
\title{\LARGE \bf
Geometry-Aware Humanoid Retargeting via Manifold ODEs$^{*}$
}
\author{Pengyu Zhu$^{1}$, Heye Wang$^{1}$, Xi Wang$^{2}$, Xuyang Wu$^{1}$, He Kong$^{1}$, and Guodong Shi$^{2}$%
\thanks{$^{1}$P. Zhu, H. Wang, X. Wu, and H. Kong are with the Southern University of Science and Technology, Shenzhen 518055, China.
Emails: \texttt{\{zhupy2025,12532826\}@mail.sustech.edu.cn}, \texttt{\{wuxy6,kongh\}@sustech.edu.cn}.}%
\thanks{$^{2}$X. Wang and G. Shi are with the University of Sydney, Sydney, Australia.
Emails: \texttt{\{xi.wang,guodong.shi\}@sydney.edu.au}.}%
\thanks{$^{*}$The first three authors contribute equally to this work with their order decided by a lucky draw, which the order of the last three authors mirrors.}%
}

\begin{document}
\maketitle
\thispagestyle{empty}
\pagestyle{empty}

\begin{abstract}
In this paper, we focus on interaction-preserving motion retargeting, which transfers human demonstrations to humanoids while maintaining spatial relationships with objects and terrain. Previous work, such as OmniRetarget, addresses this task through temporally coupled constrained optimization, requiring repeated sequential quadratic programming (SQP) iterations at each frame. We study this problem from an ordinary differential equation (ODE) perspective by modeling the continuous-time limit of its framewise Karush–Kuhn–Tucker (KKT) conditions as an ODE on the robot configuration manifold. This viewpoint offers two benefits: (i) it connects OmniRetarget’s SQP iterations to a backward discretization of the ODE, enabling us to establish their local convergence under suitable regularity assumptions; and (ii) it leads to TaSP-KKT (tangent-space parametrization for KKT flow), a new retargeting method based on a forward discretization of the KKT flow. The proposed method achieves the same first-order convergence to the ODE solution as the corresponding backward scheme while requiring only one weighted tangent-cone projection and a closed-form manifold update per step, thereby avoiding repeated inner SQP iterations. Numerical experiments show that TaSP-KKT achieves retargeting quality comparable to, though slightly lower than, that of OmniRetarget while running approximately twice as fast.  
\end{abstract}

\section{Introduction}
\label{sec:introduction}

Whole-body tasks such as carrying objects and stepping onto a platform require humanoid robots to coordinate their limbs while maintaining contact with objects and the environment~\cite{gu2026humanoid,gu2026evolution}. Human demonstrations provide examples of this coordination and can serve as references for learning robot behaviors. However, differences in body proportions, joint structure, and motion limits prevent human motions from being transferred directly to a robot. Generating usable robot motion references therefore requires adapting both the demonstrated movements and their interactions with the scene to the robot’s embodiment. Motion retargeting aims to enable a target character or robot to reproduce a source motion despite differences in body proportions or skeletal structure while preserving the motion’s essential characteristics.

Early approaches to motion retargeting include the spacetime optimization of Gleicher~\cite{gleicher1998retargetting} and the inverse-rate formulation of Choi and Ko~\cite{choi2000online}. More recently, General Motion Retargeting (GMR)~\cite{araujo2026gmr} combines local scaling with differential inverse kinematics to generate humanoid motion references. Interaction-preserving motion retargeting, which maintains spatial relationships and contacts during motion transfer, has also received much attention. For example, Contact-Aware Retargeting~\cite{villegas2021contact} combines a geometry-conditioned recurrent network with encoder-space optimization to reduce interpenetration and preserve self-contacts and ground contacts. M-R$^{2}$ET~\cite{zhang2024modular} uses a learned model to predict residual motion corrections that account for skeletal relationships and body geometry, aiming to reduce interpenetration and missing contacts when transferring motion between characters with different skeletal structures.

Among the most recent work is OmniRetarget~\cite{yang2026omniretarget}, which uses an interaction mesh to represent spatial relationships among the body, objects, and terrain. At each frame, it minimizes Laplacian mesh deformation subject to kinematic constraints using a sequential quadratic programming (SQP) solver. The resulting references reduced penetration and foot skating and supported object carrying and terrain traversal on a Unitree G1 humanoid.

In this paper, we study humanoid motion retargeting and OmniRetarget from the perspective of ordinary differential equations (ODEs). Our motivation is twofold. First, OmniRetarget solves a sequence of temporally coupled constrained optimization problems, but the original work does not establish convergence guarantees for its SQP iterations. The longstanding connections between optimization and ODEs provide tools for analyzing convergence~\cite{helmke1994optimization,schropp2000dynamical,fiori2005quasigeodesic}. These connections motivate us to establish theoretical guarantees for OmniRetarget through a dynamical formulation.

Second, an ODE formulation can guide the development of new algorithms. In deep learning, Haber et al.~\cite{haber2018learning} use a continuous-time interpretation of residual networks to develop multiscale training methods, which attains higher classification accuracy and lower training cost. In diffusion generative modeling, Lu et al.~\cite{lu2022dpmsolver} exploit the structure of diffusion ODEs to derive DPM-Solver, which use with high-order ODE schemes to accelerate sampling. These examples motivate using the structure of retargeting dynamics to design more efficient numerical methods.

ODE-based methods have also been explored in humanoid planning and retargeting. Klein et al.~\cite{klein2022riemannian} generate robot arm motions as piecewise geodesics governed by second-order ODEs under an inertia-based Riemannian metric. Their subsequent work~\cite{klein2023region} modifies this metric using barrier functions to generate collision-avoiding trajectories. Li et al.~\cite{li2026unified} use a learned conditional ODE flow to adapt motions across characters with the same skeletal topology but different body proportions. MoReFlow~\cite{kim2025moreflow} formulates human-to-robot motion mapping as an ODE flow in latent space and uses flow matching to train a neural network that parameterizes its velocity field.

In this paper, we model the evolution of the optimal robot configurations generated by OmniRetarget as an ODE on \(\mathcal M=\mathbb S^3\times\mathbb R^d\), where \(\mathbb S^3\) represents the floating-base orientation using a unit quaternion, and \(\mathbb R^d\) contains the base translation and joint coordinates. We then use this ODE formulation to analyze the convergence of OmniRetarget's SQP solver and derive a computationally efficient retargeting algorithm. Our contributions are summarized as follows:
\begin{enumerate}
 \item We develop a manifold ODE formulation of OmniRetarget. In particular, we characterize the evolution of OmniRetarget’s framewise KKT point as a projected flow on the robot configuration manifold \(\mathcal M=\mathbb S^3\times\mathbb R^d\). The formulation provides a common basis for analyzing the existing solver and designing new retargeting algorithms.

 \item We connect OmniRetarget’s SQP iterations to a backward discretization of the KKT flow. Through this connection, we interpret the SQP iterations as Newton steps for the backward discretization and establish local convergence of the SQP solver.

 \item We propose TaSP-KKT, a retargeting algorithm based on a forward discretization of the KKT flow. It achieves the same convergence order as the SQP-based backward scheme. Each step requires only one weighted tangent-cone projection and a closed-form manifold update, avoiding repeated inner SQP iterations and thus reducing the computational cost. Preliminary numerical experiments show that TaSP-KKT achieves retargeting quality comparable to that of OmniRetarget while running approximately twice as fast.
\end{enumerate}

\section{Preliminaries}
\label{sec:preliminaries}
\subsection{Ordinary Differential Equations}
\label{sec:ordinary-differential-equations}

An ODE describes the evolution
of a state $y(\tau)\in\mathbb{R}^n$ over continuous time $\tau$.
At each time $\tau$, the vector field $F(\tau,y(\tau))$ assigns
a rate of change to the current state. Together with an initial
state $y_0$, the ODE defines the initial-value problem
\begin{equation}
    \dot{y}(\tau)=F(\tau,y(\tau)),
    \qquad y(0)=y_0,
    \label{eq:prelim-ode}
\end{equation}
where $\dot{y}(\tau)=\mathrm{d}y(\tau)/\mathrm{d}\tau$.
A solution of~\eqref{eq:prelim-ode} on $[0,T]$ is a trajectory $y:[0,T] \to \mathbb R^n$ that starts at $y_0$ and whose
derivative agrees with the vector field at every time.

\paragraph{Discrete approximation}
To approximate the continuous-time trajectory numerically,
we discretize time and advance the state in finite steps.
For a step size $\eta>0$ and sampling times $\tau_k=k\eta$,
a one-step method generates approximations $y_k\approx y(\tau_k)$
according to
\begin{equation*}
    y_{k+1}=\Psi_\eta(\tau_k,y_k),
    \qquad y_0=y(0),
    \label{eq:prelim-discrete-flow}
\end{equation*}
where $\Psi_\eta$ denotes the numerical update map.
For example, the forward and backward Euler methods are
given by
\begin{equation}
    \begin{aligned}
        \text{Forward Euler:}\quad
        y_{k+1}&=y_k+\eta F(\tau_k,y_k),\\
        \text{Backward Euler:}\quad
        y_{k+1}&=y_k+\eta F(\tau_{k+1},y_{k+1}).
    \end{aligned}
    \label{eq:prelim-euler-methods}
\end{equation}
Forward Euler evaluates the vector field at the known current
state and thus provides an explicit update. Backward Euler
evaluates the vector field at the unknown next state and
therefore requires solving an implicit equation, which is
generally nonlinear~\cite{hairer2015numerical}.

\paragraph{Convergence order}
Convergence order quantifies how rapidly the numerical error
decreases as the step size $\eta$ tends to zero over a fixed
time interval $[0,T]$. Let $n_{\max}=\lfloor T/\eta\rfloor$.
A method has global convergence order $p\geq1$ if, for all
sufficiently small $\eta$,
\begin{equation*}
    \max_{0\leq k\leq n_{\max}}
    \left\lVert y_k-y(\tau_k)\right\rVert_2
    \leq C_T\eta^p,
    \label{eq:prelim-global-error}
\end{equation*}
where the constant $C_T$ may depend on $T$ but is independent of $\eta$.
Under standard smoothness assumptions, both Euler methods have first-order convergence~\cite{hairer2015numerical}.

\subsection{ODEs on Robot Configuration Space}
\label{sec:manifold-odes}

We represent the configuration of a floating-base robot as
\begin{equation}
    \begin{aligned}
        q=(r,x)\in\mathcal M&:=\mathbb S^3\times\mathbb R^d,\\
        \mathbb S^3&:=\{r\in\mathbb R^4:\|r\|_2=1\}.
    \end{aligned}
    \label{eq:prelim-configuration}
\end{equation}
The unit quaternion $r=(r_0,r_1,r_2,r_3)^\top$ represents the base orientation, while $x$ contains the base translation and joint coordinates. Quaternion signs are chosen consistently across frames. 
The tangent space $T_q\mathcal M$ consists of the velocities of smooth curves on $\mathcal M$ passing through $q$. Differentiating the constraint $r^\top r=1$ shows that a quaternion velocity $\xi_r$ must satisfy $r^\top\xi_r=0$. Thus, a tangent quaternion vector is perpendicular to $r$, and the tangent space of $\mathcal M$ at $q$ is given by
\begin{equation}
    \begin{aligned}
        T_q\mathcal M&=T_r\mathbb S^3\times\mathbb R^d,\text{ }
        T_r\mathbb S^3 =\{\xi_r\in\mathbb R^4:r^\top\xi_r=0\}.
    \end{aligned}
    \label{eq:prelim-tangent-space}
\end{equation}

An ODE on the configuration manifold must generate tangent velocities~\cite[Ch.~IV]{hairer2006geometric}:
\begin{equation}
    \dot q(\tau)=V(\tau,q(\tau)),\qquad V(\tau,q)\in T_q\mathcal M.
    \label{eq:prelim-manifold-flow}
\end{equation}
Equivalently,
\begin{equation}
    \dot r=V_r(\tau,r,x),\qquad
    \dot x=V_x(\tau,r,x),\qquad r^\top V_r=0.
    \label{eq:prelim-configuration-ode}
\end{equation}

Consider two unit quaternions $r,r'\in\mathbb{S}^3$ in the same hemisphere.
Let $v\in T_r\mathbb{S}^3$ denote the orthogonal projection
of their displacement onto the tangent space at $r$:
\begin{equation}
    v:=(I-rr^\top)(r'-r)
      =(I-rr^\top)r'.
    \label{eq:prelim-tangent-projection}
\end{equation}
The point $r'$ can therefore be decomposed as
$r'=\alpha r+v$, where $\alpha \in \mathbb R$.
Since $r^\top v=0$ and $r,r'$ have unit norm,
\begin{equation}
    1=\|r'\|_2^2=\alpha^2+\|v\|_2^2.
\end{equation}
Selecting the positive root gives
$\alpha=\sqrt{1-\|v\|_2^2}$.
Thus, for any $v\in T_r\mathbb{S}^3$ with $\|v\|_2<1$,
the tangent projection uniquely determines $r'$ within
the hemisphere through the tangent space parametrization
(TaSP) map~\cite{hairer2001manifolds}:
\begin{equation*}
    r'=\operatorname{TaSP}_r(v)
      :=\sqrt{1-\|v\|_2^2}\,r+v.
\end{equation*}

Combining this parametrization with forward Euler yields
the TaSP--Euler method for~\eqref{eq:prelim-configuration-ode}.
Starting from $q_0=q(0)$, we set
$v_k=\eta V_r(\tau_k,q_k)$ and compute $q_{k+1}=(r_{k+1},x_{k+1})$ using
\begin{equation}
\begin{aligned}
    r_{k+1}
    &=\operatorname{TaSP}_{r_k}
      \bigl(\eta V_r(\tau_k,q_k)\bigr),\\
    x_{k+1}
    &=x_k+\eta V_x(\tau_k,q_k).
\end{aligned}
\label{eq:prelim-tasp}
\end{equation}
The update preserves $q_{k+1}\in\mathcal{M}$ when
$\eta\|V_r(\tau_k,q_k)\|_2<1$. The forward TaSP--Euler method has first-order convergence~\cite{hairer2015numerical}, i.e.,
\begin{equation}
    \max_{0\leq k\leq\lfloor T/\eta\rfloor}
    \|q_k-q(\tau_k)\|_2
    \leq C_T\eta.
    \label{eq:prelim-tasp-convergence}
\end{equation}

A gradient flow for a smooth function
$f:\mathcal{M}\to\mathbb{R}$ evolves along the negative
Riemannian gradient of $f$:
\begin{equation}
    \dot{q}(\tau)=-\mathfrak{D}_q f(q(\tau)),
    \qquad q(0)=q_0,
    \label{eq:prelim-gradient-flow}
\end{equation}
where $\mathfrak{D}_q f$ denotes the Riemannian gradient, which is obtained by projecting the
Euclidean gradient onto $T_q\mathcal{M}$.
For $q=(r,x)$, the orthogonal projector is
$$\Pi_q=\operatorname{diag}(I_4-rr^\top,I_d),$$
which gives
\begin{equation}
\begin{aligned}
    \mathfrak{D}_q f
    &=\Pi_q\nabla f\\
    &=\begin{bmatrix}
        (I_4-rr^\top)\nabla_r f\\
        \nabla_x f
    \end{bmatrix}
    =\begin{bmatrix}
        \operatorname{grad}_r f\\
        \nabla_x f
    \end{bmatrix}.
\end{aligned}
\label{eq:prelim-product-gradient}
\end{equation}
Here, $\nabla f$ denotes the Euclidean gradient, and
$\operatorname{grad}_r f:=(I_4-rr^\top)\nabla_r f$
is its Riemannian gradient with respect to the quaternion
component.

\subsection{SQP-based Motion Retargeting}
\label{sec:omniretarget-problem}

OmniRetarget~\cite{yang2026omniretarget} transfers human motion to a robot by preserving the spatial relationships among the body, manipulated objects, and terrain. Consider a human demonstration over the time interval $[0,T]$, uniformly sampled with a sampling period $\delta>0$. The sampled demonstration is represented by a sequence $\{P_t\}_{t=0}^{M}$, where $P_t=[p_{t,1},\ldots,p_{t,N}]^\top\in\mathbb{R}^{N\times3}$ contains the 3D positions of human body keypoints and points sampled from object and terrain surfaces. At each sample $t$, OmniRetarget constructs an interaction mesh by applying Delaunay tetrahedralization~\cite{si2005meshing} to $P_t$. Let $L_t\in\mathbb{R}^{N\times N}$ denote the associated graph Laplacian~\cite{sorkine2004laplacian}, defined such that the $i$th row of $L_tP_t$ represents the displacement of vertex $i$ from the mean position of its neighbors~\cite{yang2026omniretarget}.

Let $q_t\in\mathcal M$ denote the retargeted robot configuration at frame $t$. The map $\Phi:\mathcal M\to\mathbb R^{N\times3}$ returns the corresponding target mesh vertices by combining robot keypoints computed through forward kinematics with prescribed object and terrain points. At each frame, OmniRetarget obtains the desired robot configuration by solving the following constrained optimization problem
\begin{equation}
    \begin{aligned}
    q_t^\star\in\arg\min_{q\in \mathcal M}\quad&
        f_t(q)+\frac{1}{2\delta}\|q-q_{t-1}\|_Q^2\\
    \text{s.t.}\quad&
        g_{t,i}(q)\leq0,\qquad i=1,\ldots,m,\\
        &h_{t,j}(q)=0,\qquad j=1,\ldots,\ell.
    \end{aligned}
    \label{eq:prelim-retargeting}
\end{equation}
The objective and constraints have the following roles:
\begin{itemize}
    \item \emph{Interaction preservation.} The loss is defined as
    \begin{equation}
        \begin{aligned}
            f_t(q)&:=\|e_t(q)\|_2^2,\\
            e_t(q)&:=\operatorname{vec}\!\left(L_t[\Phi(q)-P_t]\right),
        \end{aligned}
        \label{eq:prelim-interaction-loss}
    \end{equation}
    where $\operatorname{vec}$ stacks the columns of a matrix. The residual compares corresponding vertices' offsets from their neighbors' average positions in the robot and human interaction meshes. Minimizing $f_t$ therefore encourages preservation of local spatial relationships among the body, objects, and terrain.

    \item \emph{Temporal regularization.} With $z:=q-q_{t-1}$, define $\|z\|_Q^2:=z^\top Qz$, where $Q\in\mathbb R^{(d+4)\times(d+4)}$ is symmetric positive semidefinite. This penalty discourages large changes between consecutive frames, encouraging temporal smoothness~\cite{yang2026omniretarget}.

    \item \emph{Kinematic feasibility.} At time step $t$, the vector $g_t(q)=[g_{t,1}(q),\ldots,g_{t,m}(q)]^\top\in\mathbb{R}^m$ collects the $m$ inequality constraints enforcing non-penetration and joint position and velocity limits. Similarly, $h_t(q)=[h_{t,1}(q),\ldots,h_{t,\ell}(q)]^\top\in\mathbb{R}^{\ell}$ collects the $\ell$ equality constraints, including those specifying stance-foot positions. The subscript $t$ indicates the time dependence of these constraint functions.
\end{itemize}

Note that our temporal penalty includes an explicit factor $1/(2\delta)$, which is absent from the original formulation~\cite{yang2026omniretarget}. The two expressions differ only in their normalization and are equivalent under an appropriate rescaling of the weight matrix $Q$. OmniRetarget solves~\eqref{eq:prelim-retargeting} using SQP. At frame $t$, the iterations are initialized with the SQP solution from frame $t-1$, i.e., $\bar q_{t,0}=q_{t-1}^{\mathrm{SQP}}$. At iteration $i\geq1$, the SQP process linearizes the interaction residual and constraint functions at $\bar q_{t,i-1}$ to obtain the following trust-region subproblem for the increment $\Delta q_{t,i}\in\mathbb{R}^{d+4}$:
\begin{equation}
\begin{aligned}
\min_{\Delta q}\quad&
\bigl\|e_t(\bar q_{t,i-1})+J^e_{t,i-1}\Delta q\bigr\|_2^2\\
&+\frac{1}{2\delta}
\bigl\|\bar q_{t,i-1}+\Delta q-q_{t-1}^{\mathrm{SQP}}\bigr\|_Q^2\\
\text{s.t.}\quad&
g_{t,j}(\bar q_{t,i-1})
+J^{g_j}_{t,i-1}\Delta q\leq0,
\quad j=1,\ldots,m,\\
&
h_{t,j}(\bar q_{t,i-1})
+J^{h_j}_{t,i-1}\Delta q=0,
\quad j=1,\ldots,\ell,\\
&
\Delta q\in T_{\bar q_{t,i-1}}\mathcal M,
\qquad \|\Delta q\|_2\leq\varepsilon.
\end{aligned}
\label{eq:prelim-sqp}
\end{equation}
Here, $J^e_{t,i-1}$ denotes the Jacobian of the residual $e_t$, while $J^{g_j}_{t,i-1}$ and $J^{h_j}_{t,i-1}$ denote the row Jacobians of the scalar constraints $g_{t,j}$ and $h_{t,j}$, respectively. All Jacobians are evaluated with respect to $q$ at $\bar q_{t,i-1}$, with the frame data $P_t$ held fixed.

After solving~\eqref{eq:prelim-sqp}, we decompose the resulting increment into its quaternion and Euclidean components, $\Delta q_{t,i}=(\Delta r_{t,i},\Delta x_{t,i})$. The next iterate $\bar q_{t,i} =(\bar r_{t,i},\bar x_{t,i})$ is obtained by adding the increment to the current iterate and normalizing the quaternion component:
\begin{equation}
\begin{aligned}
\bar r_{t,i}
&=\frac{\bar r_{t,i-1}+\Delta r_{t,i}}
{\|\bar r_{t,i-1}+\Delta r_{t,i}\|_2}, \quad
\bar x_{t,i}
=\bar x_{t,i-1}+\Delta x_{t,i}.
\end{aligned}
\label{eq:prelim-sqp-update}
\end{equation}
After $N$ SQP iterations, the output for frame $t$ is defined as $q_t^{\mathrm{SQP}}:=\bar q_{t,N}$.

\section{KKT Flow of Humanoid Motion Retargeting}
\label{sec:dynamic-model}

In this section, we model the evolution of the framewise
KKT points of~\eqref{eq:prelim-retargeting} as a constrained
gradient ODE on $\mathcal{M}$ in the limit as the sampling period $\delta \to 0$. As a result, starting from a KKT point
at the initial frame, one can integrate the proposed dynamics
to track the corresponding KKT trajectory, thereby avoiding
the need to solve a separate optimization problem at each
subsequent frame. Set $\tau_t=t\delta$ and write
$f_t=f(\tau_t,\cdot)$, $g_t=g(\tau_t,\cdot)$, and
$h_t=h(\tau_t,\cdot)$.
Define the feasible set and the active inequality index set as
\[
\begin{aligned}
    \mathcal{F}_t
    &:=\{q\in\mathcal{M}:g_t(q_t)\leq0,\ h_t(q_t)=0\},\\
    \mathcal{I}(q_t)
    &:=\{i\in\{1,\ldots,m\}:g_{t,i}(q_t)=0\}.
\end{aligned}
\]

At $q\in\mathcal{F}_t$, write $\mathcal{I}=\mathcal{I}(q_t)$ and define the active constraint Jacobian
\begin{equation}
    A(q_t):=
    \begin{bmatrix}
        \mathfrak D_q g_{t,\mathcal{I}}(q_t)\\
        \mathfrak D_q h_t(q_t)
    \end{bmatrix} = \begin{bmatrix}
        \nabla_q g_{t,\mathcal{I}}(q_t)\\
        \nabla_q h_t(q_t)
    \end{bmatrix}\Pi_{q_t}.
    \label{eq:dynamic-active-jacobian}
\end{equation}

Write $\mathcal {NC}(q_t)$ and $\mathcal {TC}(q_t)$ for the outward normal and
tangent cones to $\mathcal{F}_t$ relative to $\mathcal{M}$.
At a feasible point $q$, these cones admit the
representations~\cite{bergmann2019intrinsic}
\begin{equation}
    \mathcal {NC}(q_t)=
    \left\{
        A(q_t)^\top
        \begin{bmatrix}
            \lambda_{\mathcal I}\\
            \mu
        \end{bmatrix}
        :
        \lambda\in\mathbb{R}_+^{|\mathcal{I}|},
        \ \mu\in\mathbb{R}^{\ell}
    \right\},
    \label{eq:dynamic-normal-cone}
\end{equation}
\begin{equation}
    \mathcal {TC}(q_t)=
    \left\{
        w\in T_q\mathcal{M}:
        \langle n,w\rangle\leq0
        \ \text{for all }n\in \mathcal {NC}(q_t)
    \right\}.
    \label{eq:dynamic-tangent-cone}
\end{equation}

Since $\mathcal {TC}(q_t)$ is a nonempty closed convex cone,
any matrix $Q=Q^\top\succ0$ defines a unique weighted
projection
\begin{equation}
    \mathcal{P}_{t,q}^Q(v)
    :=
    \underset{w\in \mathcal {TC}(q_t)}{\operatorname{argmin}}
    \frac12\|w-v\|_Q^2.
    \label{eq:dynamic-tangent-projection}
\end{equation}
Given any $q\in\mathcal{F}_t$ at which $A(q_t)$ has full
row rank, any $v\in\mathbb{R}^{d+4}$, any
$w\in \mathcal {TC}(q_t)$, the projection is characterized by
\begin{equation}
    \mathcal{P}_{t,q}^Q(v)=w
    \quad\Longleftrightarrow\quad
    \begin{cases}
        \Pi_qQ(v-w)\in \mathcal {NC}(q_t),\\
        \langle Q(v-w),w\rangle=0.
    \end{cases}
    \label{eq:dynamic-projection-characterization}
\end{equation}

For analysis, we make the following regularity and the linear independence constraint
qualification (LICQ) assumptions~\cite{hairer2015numerical,bergmann2019intrinsic}.
In Assumptions 1-3, $\varphi_t$ denotes any
of the scalar functions $f_t$, $g_{t,i}$ ($i=1,\ldots,m$),
or $h_{t,j}$ ($j=1,\ldots,\ell$).

\begin{assumption}[Boundedness]
\label{asm:boundedness}
There exists a constant $B>0$, independent of $t$, such that
\begin{equation}
    |\varphi_t(q)|\leq B, \qquad \forall q\in\mathcal{F}_t,
\end{equation}
holds for every frame $t$ and every function $\varphi_t$.
\end{assumption}

\begin{assumption}[Lipschitz continuity]
\label{asm:lipschitz}
There exists a constant $K>0$, independent of $t$, such
that
\[
    |\varphi_t(q)-\varphi_t(q')|
    \leq K\|q-q'\|_2,
    \qquad \forall q,q'\in\mathcal{F}_t,
\]
holds for every frame $t$ and every function $\varphi_t$.
\end{assumption}
\begin{assumption}[$L$-smoothness]
\label{asm:Lsmooth}
There exists a constant $L>0$, independent of $t$, such
that
\[
    \bigl\|
        \mathfrak{D}_q\varphi_t(q)
        -\mathfrak{D}_q\varphi_t(q')
    \bigr\|_2
    \leq L\|q-q'\|_2,
    \qquad \forall q,q'\in\mathcal{F}_t.
\]
holds for every frame $t$ and every function $\varphi_t$ (note that here the Riemannian gradients are compared as vectors
in the ambient space $\mathbb{R}^{d+4}$).
\end{assumption}

\begin{assumption}[LICQ]
\label{asm:LICQ}
For every frame $t$ and every KKT point $q_t^\star$,
the active constraint Jacobian $A(q_t^\star)$ has
full row rank.
\end{assumption}

\label{sec:dynamic-kkt}

Under these assumptions, we model the evolution of the
framewise KKT solutions as a constrained gradient ODE,
as formalized in the following theorem.

\begin{theorem}[KKT flows]
\label{lem:dynamic-kkt-flow}
Suppose Assumptions~\ref{asm:boundedness}--\ref{asm:LICQ}
hold, and
$Q=Q^\top\succ0$.
Let $q_0^\star=q_0\in\mathcal{F}$.
For each $k$, let $q_{k+1}^\star$ be a KKT point
of~\eqref{eq:prelim-retargeting} at $\tau_{k+1}$,
with reference configuration $q_k^\star$.
As $\delta\to0$,
these KKT conditions yield
\begin{equation}
\begin{aligned}
    \dot{q}^\star(\tau)
    &=\mathcal{P}_{q^\star(\tau)}^Q
      \bigl(-Q^{-1}\mathfrak{D}_q
      f(\tau,q^\star(\tau))\bigr),\\
    q^\star(\tau)&\in\mathcal{F},
    \qquad q^\star(0)=q_0.
\end{aligned}
\label{eq:dynamic-gradient-flow}
\end{equation}
\end{theorem}
\vspace{5pt}

The proof is in Appendix~\ref{app:dynamic-kkt}. Theorem~\ref{lem:dynamic-kkt-flow} suggests computing the entire retargeted motion by solving a single initial-value problem. Starting from $q_0$, integrating the projected flow over the full motion interval yields a configuration trajectory whose sampled states provide the retargeted frames. Thus, humanoid motion retargeting can be performed by integrating one ODE instead of solving a separate optimization problem at every frame.

\section{Discrete Approximation of the KKT Flow}
\label{sec:dynamic-order}
To compute the KKT trajectory numerically, we construct
discrete approximations of the KKT flow. Define the
projected velocity field
\begin{equation}
    V(\tau,q):=\mathcal{P}_q^Q
    \bigl(-Q^{-1}\mathfrak{D}_q f(\tau,q)\bigr).
    \label{eq:dynamic-descent-velocity}
\end{equation}
For a step size $\eta>0$ and a time grid $\tau_k=k\eta$,
the backward and forward TaSP--Euler updates
for~\eqref{eq:dynamic-gradient-flow} are, respectively,
\begin{equation}
    \begin{aligned}
        r_k^{\mathrm{BE}}
        &=\operatorname{TaSP}_{r_{k+1}^{\mathrm{BE}}}
        \bigl(-\eta V_r(\tau_{k+1},q_{k+1}^{\mathrm{BE}})\bigr),\\
        x_k^{\mathrm{BE}}
        &=x_{k+1}^{\mathrm{BE}}
        -\eta V_x(\tau_{k+1},q_{k+1}^{\mathrm{BE}}),
    \end{aligned}
    \label{eq:dynamic-backward-euler}
\end{equation}
and
\begin{equation}
    \begin{aligned}
        r_{k+1}^{\mathrm{FE}}
        &=\operatorname{TaSP}_{r_k^{\mathrm{FE}}}
        \bigl(\eta V_r(\tau_k,q_k^{\mathrm{FE}})\bigr),\\
        x_{k+1}^{\mathrm{FE}}
        &=x_k^{\mathrm{FE}}
        +\eta V_x(\tau_k,q_k^{\mathrm{FE}}).
    \end{aligned}
    \label{eq:dynamic-forward-euler}
\end{equation}

In this section, we first establish first-order convergence
of the backward TaSP--Euler method~\eqref{eq:dynamic-backward-euler}.
We then characterize the backward update by a nonlinear system on
$\mathcal M$. For a fixed reference configuration $q_k$, we show that
the SQP iterations~\eqref{eq:prelim-sqp} converge locally and that
their limit differs from a nearby feasible solution of the backward
system by $\mathcal O(\eta^2)$. Building on the fact that the forward TaSP--Euler method has already been proved to converge with order one~\cite{hairer1996solving,solo2021convergence}, we propose TaSP-KKT, an algorithm for OmniRetargeting based on the update in~\eqref{eq:dynamic-forward-euler}. By evaluating the velocity field at the current state and advancing the configuration explicitly, TaSP-KKT avoids the nonlinear solve and associated Newton iterations required by the backward formulation, thereby reducing the computational cost per step.

\subsection{Backward Euler Method and SQP Procedure}

We establish the convergence order of the backward TaSP--Euler method.

\begin{lemma}
\label{lem:dynamic-backward-tasp-order}
Suppose that Assumptions~\ref{asm:boundedness}--\ref{asm:LICQ}
hold, and let $q:[0,T]\to\mathcal F$ be an exact $C^2$ solution
of~\eqref{eq:dynamic-gradient-flow}.
Then, for all sufficiently small step sizes $\eta>0$, the backward
TaSP--Euler iterates initialized at $q_0^{\mathrm{BE}}=q(0)$ satisfy
the first-order global error bound
\begin{equation}
    \max_{0\le k\le n_\eta}
    \|q_k^{\mathrm{BE}}-q(\tau_k)\|
    \le C_{\mathrm{BE},T}\eta,
    \label{eq:dynamic-backward-tasp-global-error}
\end{equation}
where $n_\eta=\lfloor T/\eta\rfloor$.
\end{lemma}

The proof is provided in
Appendix~\ref{app:dynamic-backward-tasp-order}.
We next express the implicit backward TaSP update
in~\eqref{eq:dynamic-backward-euler} as an equivalent nonlinear system.

\begin{lemma}
\label{lem:dynamic-backward-system}
Suppose that Assumptions~\ref{asm:boundedness}--\ref{asm:LICQ}
hold. Let $Q=Q^\top\succ0$, $\eta>0$, and
$q_k^{\mathrm{BE}}\in\mathcal F$. Set $q=q_{k+1}^{\mathrm{BE}}=(r,x)\in\mathcal F$,
and set $\mathcal I=\mathcal I(q)$. 
If $\eta\|V_r(\tau_{k+1},q)\|<1$, then $q$ satisfies
the backward TaSP update~\eqref{eq:dynamic-backward-euler}
if and only if there exists a multiplier vector
$\pi_{k+1}=(\lambda_{\mathcal I},\mu)$ such that
\begin{equation}
\label{eq:dynamic-backward-tasp-system}
    \begin{aligned}
        &0=\Pi_qQ\Pi_q(q-q_k^{\mathrm{BE}})
            +\eta\mathfrak D_qf_{k+1}(q)
            +\eta A(q)^\top\pi_{k+1},\\
        &0=\langle\pi_{k+1},A(q)\Pi_q(q-q_k^{\mathrm{BE}})\rangle,\\
        & \lambda_{\mathcal I}\ge0, \quad q\in\mathcal F.
    \end{aligned}
\end{equation}
\end{lemma}

The proof is in Appendix~\ref{app:dynamic-backward-system}.

\begin{theorem}[SQP Convergence]
\label{thm:dynamic-order}
\label{thm:dynamic-sqp-one-step}
Fix a reference configuration $p=q_k\in\mathcal F$ and the
frame data at $\tau_{k+1}$. Let $\mathcal I=\mathcal I(p)$,
$c=(g_{\mathcal I},h)$, and $B(q):=Dc(q)\Pi_q$, with the
index set $\mathcal I$ held fixed. Set $\delta=\eta$
in~\eqref{eq:prelim-sqp}, solve each SQP subproblem exactly,
and apply the full normalization update~\eqref{eq:prelim-sqp-update}.
For sufficiently small $\eta>0$, the SQP iterates
converge to a limit
\[
    q_{k+1}^{\mathrm{SQP},\infty}(\eta)
    :=\lim_{i\to\infty}\bar q_{k+1,i};
\]
moreover, there exists a constant $C>0$, independent of
$\eta$, such that
\begin{equation}
    \bigl\|q_{k+1}^{\mathrm{SQP},\infty}(\eta)
        -q_{k+1}^{\mathrm{BE}}(\eta)\bigr\|
    \le C\eta^2.
    \label{eq:dynamic-sqp-one-step-error}
\end{equation}
\end{theorem}
\vspace{5pt}
The proof is left in Appendix~\ref{app:dynamic-order}.
Theorem~\ref{thm:dynamic-order} shows that, for a fixed reference
configuration $q_k$, the SQP iterations converge locally to a limit
within $\mathcal O(\eta^2)$ of $q_{k+1}^{\mathrm{BE}}$.
By Lemma~\ref{lem:dynamic-backward-tasp-order}, the backward iterates
approximate the KKT flow~\eqref{eq:dynamic-gradient-flow} with
$\mathcal O(\eta)$ global error. This connection provides a
KKT-flow interpretation of the SQP limit together with a local
convergence guarantee for the SQP iterations.

\subsection{Forward TaSP--Euler: The TaSP-KKT Method}

We first state the convergence order of the forward TaSP--Euler
update~\eqref{eq:dynamic-forward-euler}.

\begin{theorem}[First-order convergence of forward TaSP]
\label{thm:dynamic-forward-tasp-order}
Suppose that Assumptions~\ref{asm:boundedness}--\ref{asm:LICQ}
hold. Let $q:[0,T]\to\mathcal F$ be an exact $C^2$ solution
of~\eqref{eq:dynamic-gradient-flow}.
Then the forward TaSP--Euler iterates
in~\eqref{eq:dynamic-forward-euler}, initialized at $q_0^{\mathrm{FE}}=q(0)$, satisfy
\begin{equation}
    \max_{0\le k\le n_\eta}
    \|q_k^{\mathrm{FE}}-q(\tau_k)\|_2
    \le C_{\mathrm{FE},T}\eta,
    \label{eq:dynamic-forward-tasp-global-error}
\end{equation}
for all sufficiently small $\eta>0$.
\end{theorem}

\noindent\emph{Proof.}
See, for example,~\cite{hairer1996solving,solo2021convergence}.\hfill$\square$

Applying the forward TaSP--Euler method to the KKT flow yields
an explicit algorithm, which we call \emph{TaSP-KKT}
(tangent-space parametrization for KKT flow).
We summarize TaSP-KKT in Algorithm~\ref{alg:dynamic-tasp-kkt}.

\begin{algorithm}[t]
\caption{TaSP-KKT: Forward Integration of the KKT Flow}
\label{alg:dynamic-tasp-kkt}
\begin{algorithmic}[1]
\Require Initial state $q_0=(r_0,x_0)\in\mathcal F$,
         $Q\succ0$, step size $\eta>0$, horizon $T$
\Ensure Configurations $\{q_k\}_{k=0}^{n_\eta}$
\State $n_\eta\gets\lfloor T/\eta\rfloor$
\For{$k=0,\ldots,n_\eta-1$}
    \State $\tau_k\gets k\eta$
    \State $v_k\gets\mathcal P_{q_k}^{Q}
           \bigl(-Q^{-1}\mathfrak D_q f(\tau_k,q_k)\bigr)$
    \State Decompose $v_k=(v_{r,k},v_{x,k})$
    \State $s_k\gets\eta v_{r,k}$
    \State $r_{k+1}\gets\sqrt{1-\|s_k\|_2^2}\,r_k+s_k$
    \State $x_{k+1}\gets x_k+\eta v_{x,k}$
    \State $q_{k+1}\gets(r_{k+1},x_{k+1})$
\EndFor
\State \Return $\{q_k\}_{k=0}^{n_\eta}$
\end{algorithmic}
\end{algorithm}

TaSP-KKT uses only the current frame data and configuration.
Each integration step requires one weighted tangent-cone projection and a closed-form TaSP
update, avoiding the repeated linearizations and QP solves within
the framewise SQP loop. By
Lemma~\ref{lem:dynamic-backward-tasp-order} and
Theorem~\ref{thm:dynamic-forward-tasp-order}, forward and backward
TaSP--Euler both have first-order global accuracy under their stated
assumptions. Thus, TaSP-KKT reduces the computational work per integration step
while retaining the same convergence order.


\section{Experiments and Discussions}
\label{sec:experiments}
In this section, we validate our theoretical findings via some numerical experiments.
\begin{figure}[t]
    \centering
    \subfloat[\label{fig:exp-interaction-error}]{%
        \includegraphics[width=\columnwidth]{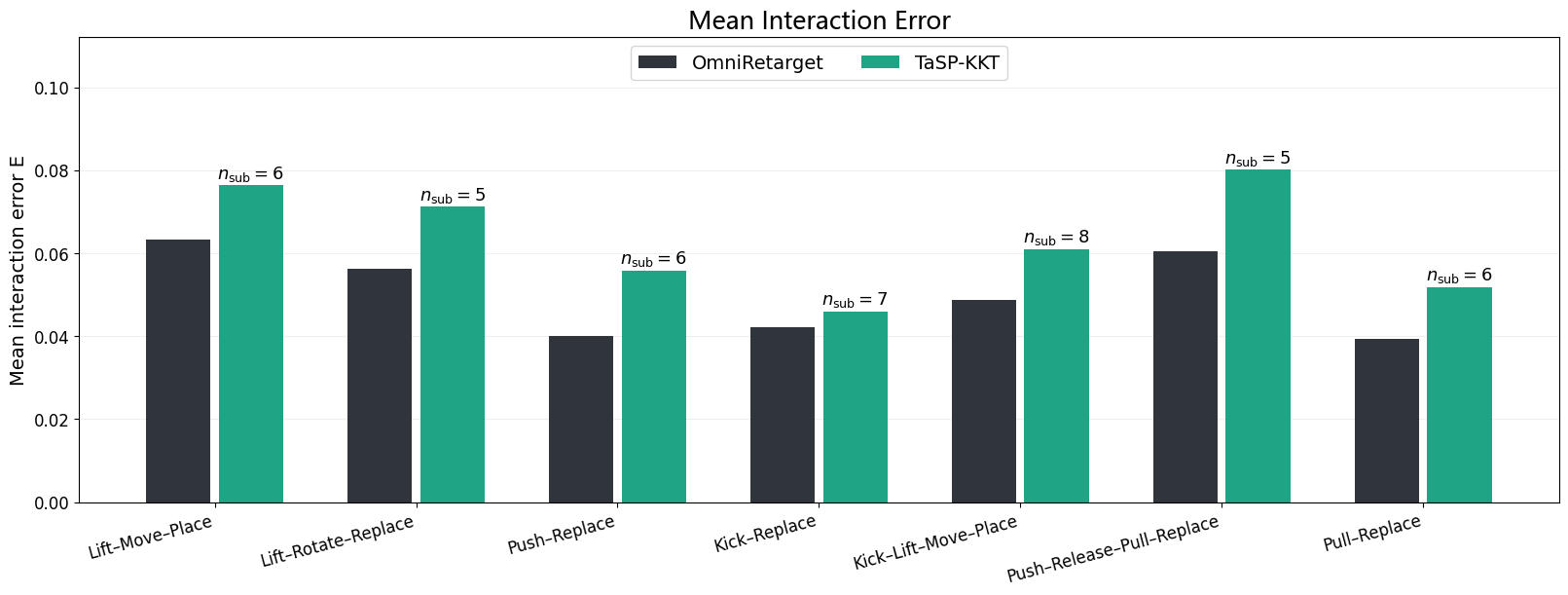}}
    \par\smallskip
    \subfloat[\label{fig:exp-solve-time}]{%
        \includegraphics[width=\columnwidth]{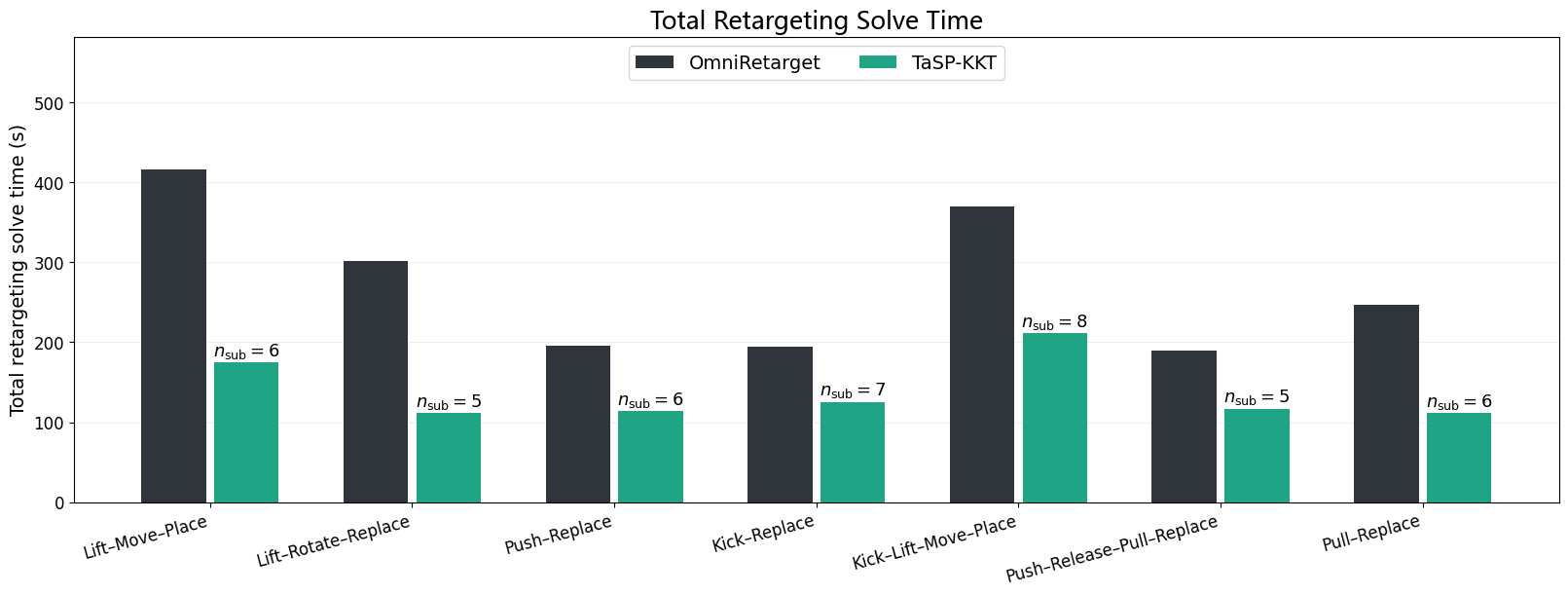}}
    \caption{OMOMO \texttt{largebox} results: (a) mean interaction loss;
    (b) total solve time.}
    \label{fig:exp-comparison}
\end{figure}

\subsection{Experimental Setup}
We evaluate TaSP-KKT for retargeting human--object interaction motions to the Unitree G1 humanoid. Human demonstrations are drawn from sequences involving the large box object (\texttt{largebox}) in the OMOMO dataset~\cite{li2023omomo}, with a sampling interval of $\delta=1/30\,\mathrm{s}$. Using the dataset's action annotations,\footnote{\url{https://github.com/lijiaman/omomo_release/blob/main/omomo_text_anno.zip}} we group the selected motions into seven patterns: Lift--Move--Place, Lift--Rotate--Replace, Push--Replace, Kick--Replace, Kick--Lift--Move--Place, Push--Release--Pull--Replace, and Pull--Replace. For each pattern, we select three complete sequences from three different subjects, yielding 21 sequences in total.

We evaluate both interaction preservation and computational efficiency.
The first metric is the mean interaction loss, computed by averaging
the framewise loss within each sequence and then taking an equally
weighted average across the three sequences in each motion pattern:
\begin{equation}
    \bar E:=\frac{1}{3}\sum_{i=1}^{3}\frac{1}{M_i}
        \sum_{t=0}^{M_i-1}f_{i,t}(q_{i,t}),
    \label{eq:exp-mean-interaction-error}
\end{equation}
where $M_i-1$ is the number of frames in sequence $i$, $q_{i,t}$ is
the retargeted robot configuration at frame $t$, and $f_{i,t}$ is
the corresponding interaction loss. A lower $\bar E$ indicates
better preservation of the reference interaction structure.

The second metric is the total retargeting solve time,
$S_{\mathrm{total}}:=\sum_{i=1}^{3}S_i$, where $S_i$ denotes
the solve time for the complete sequence $i$. Since sequence
lengths may differ across motion patterns, we compare solve
times between methods using the same sequences within each pattern.

\subsection{Results and Discussion}
As shown in Fig.~\ref{fig:exp-comparison}, TaSP-KKT requires approximately $38\%$--$63\%$ less computation time than OmniRetarget while yielding slightly higher mean interaction loss. The absolute differences in mean interaction loss are on the order of $10^{-2}$ for most motion patterns. These results illustrate our theoretical findings.

\section{Conclusion}
\label{sec:conclusion}
In this paper, we have presented an ODE perspective on interaction-preserving motion retargeting by characterizing the continuous-time limit of OmniRetarget’s framewise KKT conditions on the robot configuration manifold. This perspective supports both convergence analysis and algorithm design. By connecting the backward TaSP–Euler discretization to OmniRetarget’s SQP iterations, we establish a local convergence guarantee for the solver under suitable regularity assumptions. By discretizing the same KKT flow with a forward TaSP–Euler scheme, we derive TaSP-KKT, which achieves the same first-order convergence to the ODE solution as the backward method under the stated assumptions. Its explicit manifold update preserves the unit quaternion constraint for admissible step sizes, while replacing repeated inner SQP iterations with one weighted tangent-cone projection and a closed-form update per step. Numerical experiments demonstrate that TaSP-KKT runs approximately twice as fast as OmniRetarget while maintaining comparable retargeting quality, with a slight increase in mean interaction loss on the order of $10^{-2}$ for most motion patterns.

In future work, we will evaluate TaSP-KKT on a broader range of datasets and assess its effectiveness in downstream tasks, such as humanoid control based on reinforcement learning.

\appendices
\section{Proof of Theorem~\ref{lem:dynamic-kkt-flow}}
\label{app:dynamic-kkt}

The manifold KKT conditions~\cite{bergmann2019intrinsic}
retain the usual primal feasibility, dual feasibility,
and complementary slackness conditions. Stationarity is
expressed in the tangent space using Riemannian gradients.
For the fixed reference configuration $q_k^\star$, the
temporal penalty has Euclidean and Riemannian gradients
\[
\begin{aligned}
    \nabla_q
    \left[\frac{1}{2\delta}\|q-q_k^\star\|_Q^2\right]
    &=\delta^{-1}Q(q-q_k^\star),\\
    \mathfrak{D}_q
    \left[\frac{1}{2\delta}\|q-q_k^\star\|_Q^2\right]
    &=\delta^{-1}\Pi_qQ(q-q_k^\star).
\end{aligned}
\]
Evaluating these expressions at $q_{k+1}^\star$ gives
the framewise KKT conditions
\begin{equation}
\begin{aligned}
    0={}&\delta^{-1}\Pi_{q_{k+1}^\star}
         Q(q_{k+1}^\star-q_k^\star)\\
        &+\mathfrak{D}_q f(\tau_{k+1},q_{k+1}^\star)
         +A(q_{k+1}^\star)^\top\nu_{k+1}^\star,\\
    &q_{k+1}^\star\in\mathcal{F},
     \qquad \lambda_{k+1}^\star\geq0,\\
    &\lambda_{k+1,i}^\star g_i(q_{k+1}^\star)=0,
     \quad i=1,\ldots,m.
\end{aligned}
\label{eq:dynamic-weighted-kkt}
\end{equation}
Here, $\nu_{k+1}^\star$ stacks the active inequality
multipliers and the equality multipliers. Taking $\delta\to0$ and $\tau_k\to\tau$, we have
\[
    \begin{aligned}
        q_k^\star,\ q_{k+1}^\star
        &\longrightarrow q^\star(\tau), \text{ } \frac{q_{k+1}^\star-q_k^\star}{\delta}
        \longrightarrow\dot q^\star(\tau),\\
        (\lambda_{k+1}^\star,\mu_{k+1}^\star)
        &\longrightarrow(\lambda^\star(\tau),\mu^\star(\tau)).
    \end{aligned}
\]
Therefore, we have
\begin{equation}
    \begin{aligned}
        0={}&\Pi_{q^\star(\tau)}Q\dot q^\star(\tau)
        +\mathfrak D_qf(\tau,q^\star(\tau))+A(q^\star(\tau))^\top\nu^\star(\tau).
    \end{aligned}
    \label{eq:dynamic-limiting-kkt}
\end{equation}

Set $n^\star(\tau):=A(q^\star(\tau))^\top\nu^\star(\tau)\in \mathcal{NC}(q^\star(\tau))$.
Since the trajectory $q^\star(\tau)$ is feasible, both the forward and backward tangent directions, $\dot q^\star(\tau)$ and $-\dot q^\star(\tau)$, must belong to $\mathcal{TC}(q^\star(\tau))$. The polar-cone definition therefore gives
\[
    \langle n^\star(\tau),\dot q^\star(\tau)\rangle=0.
\]
For every $w\in \mathcal{TC}(q^\star(\tau))$, both $w$ and $\dot q^\star$ are tangent to $\mathcal M$. Hence~\eqref{eq:dynamic-limiting-kkt} implies
\[
    \begin{aligned}
        \langle &\Pi_{q^\star} Q\dot q^\star+\mathfrak D_qf,\,
                    w-\dot q^\star\rangle=\langle Q\dot q^\star+\mathfrak D_qf,\,
                    w-\dot q^\star\rangle\\
        &\quad=-\langle n^\star,w-\dot q^\star\rangle
              =-\langle n^\star,w\rangle\ge0.
    \end{aligned}
\]
This is the first-order optimality condition for minimizing
$\tfrac12w^\top Qw+\langle\mathfrak D_qf,w\rangle$
over the tangent cone. Since $Q\succ0$, the minimizer is unique. Thus,~\eqref{eq:dynamic-limiting-kkt} yields
\begin{equation}
    \dot q^\star(\tau)
    =\mathcal P_{q^\star(\tau)}^Q
      \bigl(-Q^{-1}\mathfrak D_qf(\tau,q^\star(\tau))\bigr).
    \label{eq:dynamic-projected-flow} 
\end{equation} 
The proof is completed.
\hfill$\square$

\section{Proof of Lemma~\ref{lem:dynamic-backward-tasp-order}}
\label{app:dynamic-backward-tasp-order}
\noindent\emph{Proof.}
For brevity, write $q_k=q_k^{\mathrm{BE}}$ and
$V_{k+1}:=V(\tau_{k+1},q_{k+1})$.
The boundedness and Lipschitz continuity of $f_t$, $g_t$,
and $h_t$ imply that there exist constants $M>0$ and
$L_V>0$, independent of $\tau$ and $\eta$, such that
\[
    \begin{aligned}
        \|V(\tau,u)\|&\le M,\\
        \|V(\tau,u)-V(\tau,v)\|&\le L_V\|u-v\|.
    \end{aligned}
\]
The backward TaSP update~\eqref{eq:dynamic-backward-euler}
gives
\[
    \begin{aligned}
        r_k&=\sqrt{1-\eta^2\|V_{r,k+1}\|^2}\,r_{k+1}
            -\eta V_{r,k+1},\\
        x_k&=x_{k+1}-\eta V_{x,k+1}.
    \end{aligned}
\]
Consequently,
\[
    q_{k+1}=q_k+\eta V_{k+1}+R_{k+1},
\]
where
\[
    R_{k+1}:=
    \begin{pmatrix}
        \bigl(1-\sqrt{1-\eta^2\|V_{r,k+1}\|^2}\bigr)r_{k+1}\\
        0
    \end{pmatrix}.
\]
For $0<\eta\le(2M)^{-1}$, rationalizing the square root
yields
\[
    \begin{aligned}
        \|R_{k+1}\|
        &=\frac{\eta^2\|V_{r,k+1}\|^2}
        {1+\sqrt{1-\eta^2\|V_{r,k+1}\|^2}}\le M^2\eta^2.
    \end{aligned}
\]

We introduce the frozen-field implicit sequence
\[
    \bar q_0=q_0,\qquad
    \bar q_{k+1}
    =\bar q_k+\eta V(\tau_{k+1},q_{k+1}).
\]
This iteration has the form of the classical backward Euler method, with the velocity evaluated at the backward TaSP
iterate $q_{k+1}$.
Subtracting the two recursions and summing gives $q_k-\bar q_k=\sum_{j=1}^{k}R_j.$
Hence, for $k\eta\le T$,
\[
    \|q_k-\bar q_k\|
    \le kM^2\eta^2
    \le M^2T\eta.
\]
Next, we interpolate the auxiliary sequence linearly, i.e.,
\[
    \bar q(s)=\bar q_{j-1}+(s-\tau_{j-1})V_j,
    \qquad s\in[\tau_{j-1},\tau_j].
\]
The velocity bound implies
$\|\bar q_j-\bar q(s)\|\le M\eta$.
Therefore,
\[
    \begin{aligned}
        \|q_j-\bar q(s)\|
        &\le\|q_j-\bar q_j\|+\|\bar q_j-\bar q(s)\|\\
        &\le(M^2T+M)\eta.
    \end{aligned}
\]
Let $e(s):=\|\bar q(s)-q(s)\|$.
Since $\dot q=V(\tau,q)$ and $\|\dot q\|\le M$,
\[
    \|q(\tau_j)-q(s)\|\le M\eta.
\]
It follows that $\|q_j-q(\tau_j)\|
    \le e(s)+(M^2T+2M)\eta.$
Since $q\in C^2([0,T])$, the mean value estimate
gives
\[
    \|\dot q(\tau_j)-\dot q(s)\|
    \le\eta\max_{0\le\tau\le T}\|\ddot q(\tau)\|
    =\mathcal O(\eta),
\]
uniformly in $j$ and $s\in[\tau_{j-1},\tau_j]$.
Thus, for $s\in[\tau_{j-1},\tau_j]$,
\[
    \begin{aligned}
        &\|V_j-V(s,q(s))\|\\
        &\quad\le
        \|V(\tau_j,q_j)-V(\tau_j,q(\tau_j))\|\\
        &\qquad+\|\dot q(\tau_j)-\dot q(s)\|\\
        &\quad\le L_V\|q_j-q(\tau_j)\|+\mathcal O(\eta)\\
        &\quad\le L_Ve(s)+C_*\eta,
    \end{aligned}
\]
where $C_*>0$ is independent of $\eta$, $j$, and $s$.

Because $\bar q(0)=q(0)$, the integral representations
of $\bar q$ and $q$ give
\[
    e(\tau)\le L_V\int_0^\tau e(s)\,\mathrm ds
        +C_*\tau\eta,
    \qquad 0\le\tau\le\tau_{n_\eta}.
\]
Gronwall's inequality therefore yields
\[
    e(\tau)\le C_*\tau e^{L_V\tau}\eta.
\]
At the grid points,
\[
    \begin{aligned}
        \|q_k-q(\tau_k)\|
        &\le\|q_k-\bar q_k\|+\|\bar q_k-q(\tau_k)\|\\
        &\le\bigl(M^2T+C_*T e^{L_VT}\bigr)\eta.
    \end{aligned}
\]
Thus~\eqref{eq:dynamic-backward-tasp-global-error} holds with
\[
    C_{\mathrm{BE},T}:= M^2T+C_*T e^{L_VT},
\]
which is independent of $\eta$.
\hfill$\square$

\section{Proof of Lemma~\ref{lem:dynamic-backward-system}}
\label{app:dynamic-backward-system}
\noindent\emph{Proof.}
Set $V_{k+1}:=V(\tau_{k+1},q)$, $d_{k+1}=\Pi_q(q-q_k^{\mathrm{BE}})$. TaSP inverts tangent projection, so
\[
    \eqref{eq:dynamic-backward-euler}
    \quad\Longleftrightarrow\quad
    d_{k+1}=\eta V_{k+1}.
\]
Applying~\eqref{eq:dynamic-projection-characterization} with
$u=-Q^{-1}\mathfrak D_qf_{k+1}(q)$ and $w=d_{k+1}/\eta$, there exist $\pi_{k+1,\mathcal I}\ge0$ such that 
\[
\begin{aligned}
    &-\mathfrak D_q f_{k+1}(q)
      -\eta^{-1}\Pi_q Qd_{k+1} =A(q)^\top\pi_{k+1}\in \mathcal {NC}(q_t),\\
    &\langle Q(u-w),w\rangle=\eta^{-1}
      \langle\pi_{k+1},
      A(q)d_{k+1}\rangle=0.
\end{aligned}
\]
Together with $q\in\mathcal F$, these relations complete the proof.
\hfill$\square$

\section{Proof of Theorem~\ref{thm:dynamic-order}}
\label{app:dynamic-order}

\noindent\emph{Proof.}
Fix the reference configuration $p=q_k$. Set $J_{i-1}=J^e_{k,i-1}$ and $\bar{q}_{i-1} = \bar{q}_{k,i-1}$. With the trust-region constraint omitted,
the SQP increment $\Delta_q$ at iteration $i-1$ solves
\begin{equation}
\begin{aligned}
\min_{\Delta_q\in T_q\mathcal M}\quad&
    \eta\|e(\bar{q}_{i-1})+J_{i-1}\Delta_q\|_2^2
    +\frac12\|\bar q_{i-1}+\Delta_q-p\|_Q^2\\
\text{s.t.}\quad&
    g(\bar{q}_{i-1})+J^{g}_{t,i-1}\Delta_q\le0,\\
&   h(\bar{q}_{i-1})+J^{h}_{t,i-1}\Delta_q=0.
\end{aligned}
\label{eq:sqp-contraction-subproblem}
\end{equation}
Define $H_{i-1}:=Q+2\eta J_{i-1}^\top J_{i-1}, \text{ }F_{i-1}:=2 J_{i-1}^\top e(\bar q_{i-1})$.
The KKT conditions in $\mathcal M$ are
\begin{equation}\label{eq:sqp-contraction-kkt}\begin{aligned}
0={}&
\Pi_{\bar q_{i-1}} (H_{i-1}\Delta_q
+ Q(\bar q_{i-1}-p)
+\eta F_{i-1})\\
&+\Pi_{\bar q_{t-1}}(J^g_{t,i-1})^\top\lambda
+\Pi_{\bar q_{t-1}}(J^h_{t,i-1})^\top\mu,\\[2pt]
&\Delta_q\in T_q\mathcal M,\\
&g(\bar q_{i-1})+J^g_{t,i-1}\Delta_q\le0,\\
&h(\bar q_{i-1})+J^h_{t,i-1}\Delta_q=0,\\
&\lambda\in\mathbb R_+^m,\qquad
\mu\in\mathbb R^\ell,\\
&\lambda_j
\left[g_j(\bar q_{i-1})
+J^{g_j}_{t,i-1}\Delta_q\right]=0,
\qquad j=1,\ldots,m.
\end{aligned}
\end{equation}
Therefore, the SQP iteration is the composition
\[
\bar q_{i-1}\mapsto\Delta_q
\mapsto \bar q_{i} :=\mathcal {U}(\bar q_{i-1}):=(\frac{\bar r_{i-1}+\Delta r}{\|\bar r_{i-1}+\Delta r\|}, \bar x_{i-1}+\Delta x).
\]
Now we consider the trust region bound $\epsilon$, we let $\epsilon = R \delta$ and define the neighborhood \[\mathcal D_\eta :=\{q\in\mathcal M:\|q-p\|\le \epsilon\}.\] We show that, for sufficiently small $\eta$, the map $\mathcal {U}_\eta$ is a local contraction on the $\mathcal D_\eta$, i.e.,
\begin{equation}
\|\mathcal {U}(q)-\mathcal {U}(v)\|
\le \beta \|q-v\|, \quad \mathcal U(\mathcal D_\eta)\subseteq\mathcal D_\eta. \quad \beta <1
\label{eq:sqp-contraction-map}
\end{equation} 
We omit the proof due to space constraints. By the contraction property of $\mathcal U$ on $\mathcal D_\eta$, the iteration
converges to a fixed point $\bar q_\infty$ at which $\Delta q=0$. Hence, the KKT condition become
\[
\begin{aligned}
0&=\Pi_{\bar q_\infty} Q(\bar q_\infty-p)+ \Pi_{\bar q_\infty}  \eta F_{\bar q_\infty} \\
&+\Pi_{\bar q_{t-1}}(J^g_{t,i-1})^\top\lambda
+\Pi_{\bar q_{t-1}}(J^h_{t,i-1})^\top\mu.
\end{aligned}
\]
Given that
\begin{align*}
    & \Pi_{\bar q_\infty}F_{\bar q_\infty}= \Pi_{\bar q_\infty}\nabla_q f_{k+1}({\bar q_\infty})= \mathfrak Df_{k+1}({\bar q_\infty}),  \\
    & \Pi_{\bar q_{t-1}}(J^g_{t,i-1}) = Dg_{k+1}({\bar q_\infty}), \Pi_{\bar q_{t-1}}(J^h_{t,i-1}) = Dh_{k+1}({\bar q_\infty}),
\end{align*}
therefore, with muliter \(\pi_\infty = (\lambda_{\mathcal I_\infty} , \mu_\infty),\) we have
\[
\begin{aligned}
0={}&
\Pi_{\bar q_\infty}Q(\bar q_\infty-p)
+\eta\mathfrak D_qf_{k+1}(\bar q_\infty)
+A(\bar q_\infty)^\top\pi_\infty,\\
&\lambda_{\mathcal I_\infty}\ge0,
\qquad \bar q_\infty\in\mathcal F.
\end{aligned}
\]
Compared with~\eqref{eq:dynamic-backward-tasp-system}, this system use $\Pi_{\bar q_\infty}Q$ instead of  $\Pi_qQ\Pi_q$.
For a backward solution $u=q_{k+1}^{\mathrm{BE}}$ with $u-p=\mathcal O(\eta)$, its stationarity
residual in the above system is
\[
b_\eta=\Pi_uQ(I-\Pi_u)(u-p).
\]
The unit-quaternion identity gives
\[
\|b_\eta\|
\le \frac{\|Q\|}{2}\|r_u-r_p\|^2
=\mathcal O(\eta^2).
\]
The local KKT inverse estimate therefore yields
\[
\|\bar q_\infty -q_{k+1}^{\mathrm{BE}}\|
\le C\|b_\eta\|
=\mathcal O(\eta^2).
\]

\bibliographystyle{IEEEtran-no-url}
\bibliography{retargeting-references}

\end{document}